\pdfoutput=1

\documentclass[11pt]{article}

\usepackage[final]{acl}

\usepackage{times}
\usepackage{latexsym}

\usepackage[T1]{fontenc}

\usepackage[utf8]{inputenc}

\usepackage{microtype}

\usepackage{inconsolata}

\usepackage{graphicx}

\usepackage{booktabs}
\usepackage{amsmath}
\usepackage{amssymb}
\usepackage{tabularx}
\usepackage{makecell}
\usepackage{multirow}
\usepackage{floatrow}
\usepackage{colortbl}

\title{Let’s Simplify Step by Step: Guiding LLM Towards Multilingual Unsupervised Proficiency-Controlled Sentence Simplification}

\author{Jingshen Zhang\textsuperscript{1,2}, Xin Ying Qiu\textsuperscript{1}\thanks{Corresponding author: xy.qiu@foxmail.com}, Lifang Lu\textsuperscript{1}, Zhuhua Huang\textsuperscript{1}, \\{\bf Yutao Hu\textsuperscript{1}, Yuechang Wu\textsuperscript{1}, JunYu Lu\textsuperscript{3}}\\
	\textsuperscript{1}Department of Computer Science, School of Information Science and Technology, \\
	Guangdong University of Foreign Studies, Guangzhou, China \\
	\textsuperscript{2}College of Intelligence and Computing, Tianjin University, Tianjin, China\\
	\textsuperscript{3}Lionrock AI Lab, China Merchants Research Institute of Advanced Technology, China \\
	{\tt audbut0702@163.com, xy.qiu@foxmail.com} \\
}

\begin{document}
\maketitle
\begin{abstract}
Large language models demonstrate limited capability in proficiency-controlled sentence simplification, particularly when simplifying across large readability levels. We propose a framework that decomposes complex simplifications into manageable steps through dynamic path planning,  semantic-aware exemplar selection, and  chain-of-thought generation with conversation history for coherent reasoning. Evaluation on five languages across two benchmarks shows our approach improves simplification effectiveness while reducing computational steps by 22-42\%. Human evaluation confirms the fundamental trade-off between simplification effectiveness and meaning preservation. Notably, even human annotators struggle to agree on semantic preservation judgments, highlighting the inherent complexity of this task. Our work shows that while step-by-step simplification improves control, preserving semantic fidelity during extensive simplification remains an open challenge.
\end{abstract}

\section{Introduction}
Controlled text simplification, introduced by Scarton and Specia \shortcite{scarton2018learning}, aims to automatically rewrite sentences to match the comprehension level of a specific target audience. This capability is essential for creating educational materials for children (\citealp{javourey2022simplification}), adapting  information for broader audiences (\citealp{grabar2018clear}), and ensuring content accessibility across languages (\citealp{ryan2023revisiting}).  

The field of controllable text simplificationen compasses diverse control mechanisms and methodological approaches. Control strategies can be categorized as either binary distinctions between complex and simple text \cite{blinova-etal-2023-simsum} or ordinal scales based on specific criteria such as age of acquisition \cite{oshika2024simplifyingtranslationschildreniterative}, simplicity rank \cite{paetzold2015lexenstein, glavas2015simplifying, qiang2021lsbert}, school grade level \cite{scarton2018learning, agrawal2023controlling, cripwell-etal-2023-document, mo-hu-2024-expertease}, readability scores \cite{Imperial2023flesch, farajidizaji2024possible}, or CEFR proficiency levels \cite{Imperial2023flesch,barayan2025analysing}. The scope of simplification varies across granularities, ranging from document-level \cite{cripwell-etal-2023-document, blinova-etal-2023-simsum, mo-hu-2024-expertease} and paragraph-level \cite{Imperial2023flesch, farajidizaji2024possible} to sentence-level \cite{scarton2018learning, agrawal2023controlling, barayan2025analysing} and lexical-level transformations \cite{glavas2015simplifying, paetzold2015lexenstein, qiang2021lsbert, oshika2024simplifyingtranslationschildreniterative}. Methodologically, the field has evolved from early approaches based on machine translation and machine learning \cite{paetzold2015lexenstein, scarton2018learning} to pre-trained language model (PLM) based methods \cite{qiang2021lsbert, cripwell-etal-2023-document, blinova-etal-2023-simsum, agrawal2023controlling} and, more recently, large language model (LLM) based approaches \cite{Imperial2023flesch, farajidizaji2024possible, mo-hu-2024-expertease, oshika2024simplifyingtranslationschildreniterative, barayan2025analysing}. The present study focuses specifically on CEFR proficiency-controlled sentence simplification using LLM-based methods.

CEFR proficiency-level sentence simplification represents controllable simplification aligned with the Common European Framework of Reference for Languages, a widely recognized framework for describing language ability across six levels (A1 to C2). Advances in this area include the development of specialized corpora such as the CEFR-based Sentence Profile (CEFR-SP) corpus \cite{arase2022cefr} and multilingual datasets such as README++ \cite{naous2024readme} and UniversalCEFR \cite{imperial2025universalcefrenablingopenmultilingual}.

Recent studies have revealed notable limitations in LLMs' ability to perform proficiency controlled complexity adjustments. For example, \citealp{barayan2025analysing} demonstrated success in moderate simplification tasks. However, LLMs often struggle with more challenging scenarios that require larger readability spans, such as simplifying from C2 to A1 level. Additionally, our pilot studies revealed a trade-off between reaching the target simplification level and preserving semantic meaning.

To address these challenges, we propose decomposing complex simplification into manageable incremental steps. Building on Chain-of-Thought reasoning (\citealp{wei2023chain}), we develop a novel framework that combines systematic path planning with semantic-aware exemplar selection. We evaluate this approach through comprehensive multilingual experiments across 5 languages with extensive human assessment. Our code is available at: \url{https://github.com/JasonZhang0702/simplify-step-by-step}

The key contributions of our work are the following:\\
\textbf{A dynamic programming approach for optimal path planning} that decomposes large-span simplifications into intermediate steps, improving both target level achievement (by up to 20 percentage points) and computational efficiency (22-42\% reduction in inference steps)\\
\textbf{A semantic-aware Chain-of-Thought framework} that combines exemplar selection based on meaning preservation with conversation history tracking, enabling coherent multi-step simplification while maintaining semantic fidelity.\\
\textbf{Comprehensive multilingual evaluation} across 5 languages with extensive human assessment, revealing a fundamental trade-off between readability control and meaning preservation that intensifies with simplification span—a challenge confirmed by both automatic metrics and human experts.

\section{Related Research}

\textbf{Controllable text simplification}: 
Controllable text simplification aims to adapt text complexity to specific target audiences or proficiency levels. As noted in our introduction, control objectives vary from binary classifications to fine-grained ordinal scales. Controllable approaches include explicit simplification operations \cite{alva2017learning, dong2019editnts}, attribute based control \cite{martin2020controllable}, paraphrasing \cite{maddela2021controllable} or control tokens prepended to input sequences to pre-trained language models \cite{agrawal2023controlling}.

For CEFR-controlled generation specifically, Stowe et al. \shortcite{stowe2022controlled} explored controlled generation using a concept2seq framework with CEFR and semantic role labels as control features, though they limited generation to only A1 or C2 extremes. Imperial and Madabushi \shortcite{Imperial2023flesch} examined proficiency alignment of instruction-tuned LLMs for passage-level simplification, finding that more information in prompts yields better results. Barayan et al. \shortcite{barayan2025analysing} analyzed CEFR-controlled sentence simplification across eight LLMs, demonstrating that models struggle with lower readability targets. Our paper particularly addresses the challenges of simplifying across multiple proficiency levels, where direct single-step simplification may result in insufficient simplification.

\textbf{Planning-Based Approaches in Text Generation}:
Planning mechanisms have emerged as effective strategies for improving controlled text generation across various tasks. For text summarization, hierarchical planning decomposes generation into content selection and surface realization stages \cite{moryossef2019step}. Puduppully et al. \shortcite{puduppully2019data} proposed a neural data-to-text generation model that explicitly incorporates content selection and planning stages. 
For text simplification specifically, Cripwell et al. \shortcite{cripwell-etal-2023-document} proposed a document planning approach for document-level simplification, where an LLM first generates a high-level plan outlining the document structure before performing sentence-level simplifications. They showed that planning significantly outperforms end-to-end document simplification models. Related work on sentence deletion \cite{zhong2020discourse, zhang2022predicting} uses discourse structure features from surrounding sentences to identify deletion candidates. In our work, we apply dynamic path planning to determine optimal intermediate proficiency levels between source and target, for more efficient controlled in simplification across large span of levels.

\textbf{Few-Shot Chain-of-Thought Generation}
Few-shot learning with carefully selected exemplars has proven effective for text generation tasks. Liu et al. \shortcite{liu2022what} showed that selecting in-context examples (relative to random sampling) that are semantically-similar to a test sample improves generation quality for Large Language Model. Malik et al.\shortcite{malik2024tarzantolkiencontrollinglanguage} demonstrated that selecting exemplars based on minimizing control error improves target level achievement in CEFR-controlled generation. While we control simplification accuracy with path-planning, our semantic-aware exemplar selection method prompts LLMs with best sentence pairs from validation stage for improved meaning preservation in simplification generation.

Chain-of-Thought (CoT) prompting \cite{wei2023chain} has emerged as a powerful technique for eliciting step-by-step reasoning in LLMs across various tasks. Kojima et al.\shortcite{kojima2022large} showed that even zero-shot CoT with simple prompts like ``Let's think step by step'' can improve reasoning. Zhou et al.\shortcite{zhou2023least} introduced least-to-most prompting with a top-down decomposition of the problem and a bottom-up resolution generation. The idea is to break down a complex problem into a series of simpler subproblems and then solve them in sequence. Solving each subproblem is facilitated by the answers to previously solved subproblems.

Our approach differs in two key aspects: (1) our CoT is structurally guided by the path planning algorithm, and (2) our exemplars are automatically selected based on semantic relevance from validation stage. Furthermore, we explore a chat history-based CoT design where previous simplification steps are retained in the conversation context, making the model explicitly aware of its reasoning trajectory.

\section{Methodology}

\begin{figure*}[ht]
\centering
\includegraphics[width=\linewidth]{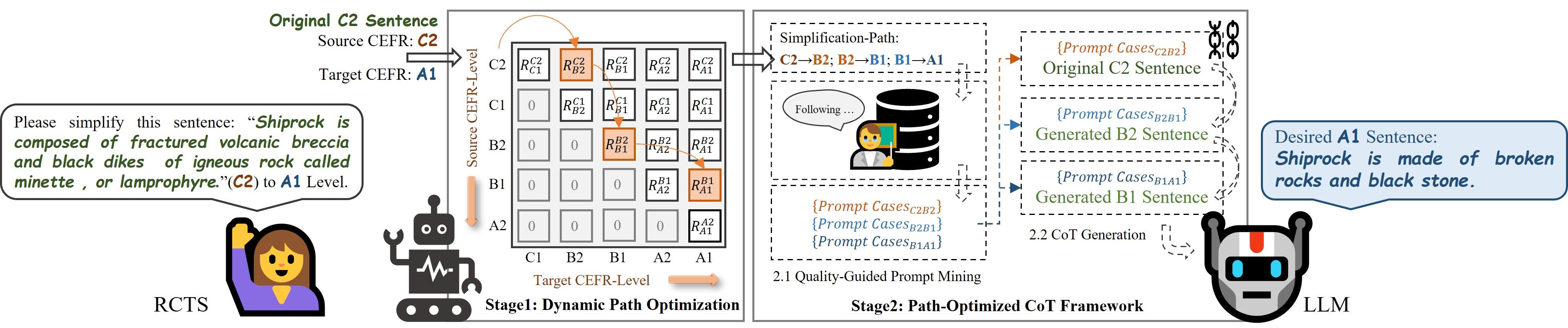}
\caption{Research Framework}
\label{fig:framework}
\end{figure*}

\subsection{Dynamic Path Planning}
Given training, validation, and test sets consisting of single sentences annotated with CEFR levels, we use an LLM to simplify sentences  from higher levels to lower levels. Let $L = \{C2, C1, B2, B1, A2, A1\}$ be the set of CEFR levels; $C_{src} \in \{C2, C1, B2\}$ be the source CEFR level; $C_{tgt} \in \{B1, A2, A1\}$ be the target CEFR level. While \citet{barayan2025analysing} use level-labeled sentences from the training set as prompts to simplify test sentences directly, we apply this approach to validation sentences first. This intermediate step allows us to construct the reward matrix and optimize the simplification path. We use a sentence-level estimator $\phi(\cdot)$ to verify if the generated sentences achieve their target levels. Based on the verification results, we populate the reward matrix $R$ where $R_{i,j}$ represents the effectiveness of simplifying from level $i$ to $j$:

\begin{equation}
R_{i,j} = \begin{cases}
    +1, & \text{if achieved level matches target} \\
    +0.5, & \text{if differs by one step} \\ 
    -1, & \text{otherwise}
\end{cases}
\end{equation}

We further normalize the reward scores to the range [0, 1] using min-max normalization. Given this reward matrix, for any source-target level pair, we formulate the path optimization problem as finding the optimal sequence $P^* = [l_0, l_1, \ldots, l_k]$ that maximizes:
\begin{equation}
P^* = {argmax}_P\sum_{i=1}^{k}R_{l_{i-1}, l_i}
\end{equation}
subject to  $l_0 = C_{src}$, $l_k = C_{tgt}$, $Level(l_i) < Level(l_{i-1})$.

This optimization is solved using dynamic programming by defining $V(i,t)$ as the maximum reward achievable in $t$ steps ending at level $i$:
\begin{equation}
V(i,t) = \max_{j < i}\{V(j,t-1) + R_{j,i}\}
\end{equation}
The optimal path can then be reconstructed by backtracking through the dynamic programming table.

\subsection{Semantic-Guided Exemplar Selection}
We begin with the validation set $D_{val} = \{(x_i, l_i)\}_{i=1}^N$, where $x_i$ represents a source sentence in $D_{val}$ with its CEFR level $l_i$ in $C_{src}$, and $N$ the total number of such sentences in $D_{val}$. We construct exemplar set for each transition step as follows:
 For each transition $(l_s, l_t)$ in optimal paths (where $l_t$ may be multiple levels below $l_s$):
\begin{itemize}
\item Generate outputs by instructing the LLM to simplify sentences at level $l_s$ to level $l_t$, with level-labeled sentences from the training set as prompt examples
\item Verify the achieved levels using sentence-level estimator $\phi(\cdot)$
\item Keep only the successful simplification pairs with exact match:
		\begin{multline}
		\hat{Y}_{s,t} = \{(x_i, \hat{y}_i) | x_i \in D_{val}, \\ l_i = l_s, \phi(\hat{y}_i) = l_t\}
		\end{multline}
		where $\hat{y}_i$ is the LLM-generated simplification of $x_i$.
\end{itemize}
For each transition $(l_s, l_t)$, we construct the exemplar set $E_{s, t}$ by selecting top $k$ sentence pairs from $\hat{Y}_{s,t}$ with the highest semantic preservation, measured with Semantic Textual Similarity (STS) (\citealp{reimers2019sentencebert}) between the original sentence $x_i$ and its simplified version $\hat{y}_i$: 
	\begin{equation}
	E_{s,t} = \{(x_i, \hat{y}_i) \in \hat{Y}_{s,t} \mid \text{rank}_{STS}(x_i, \hat{y}_i) \leq k\}
	\end{equation}
	In implementation, we set $k = 3$. If $\hat{Y}_{s,t}$ contains fewer than 3 pairs, we use all available pairs as examples. In rare cases when $\hat{Y}_{s,t}$ is empty, we select as examples single sentences from the training set labeled with $l_t$.

\subsection{Few-shot CoT with Chat History}
\begin{figure*}[h]
\centering
\includegraphics[scale=0.38]{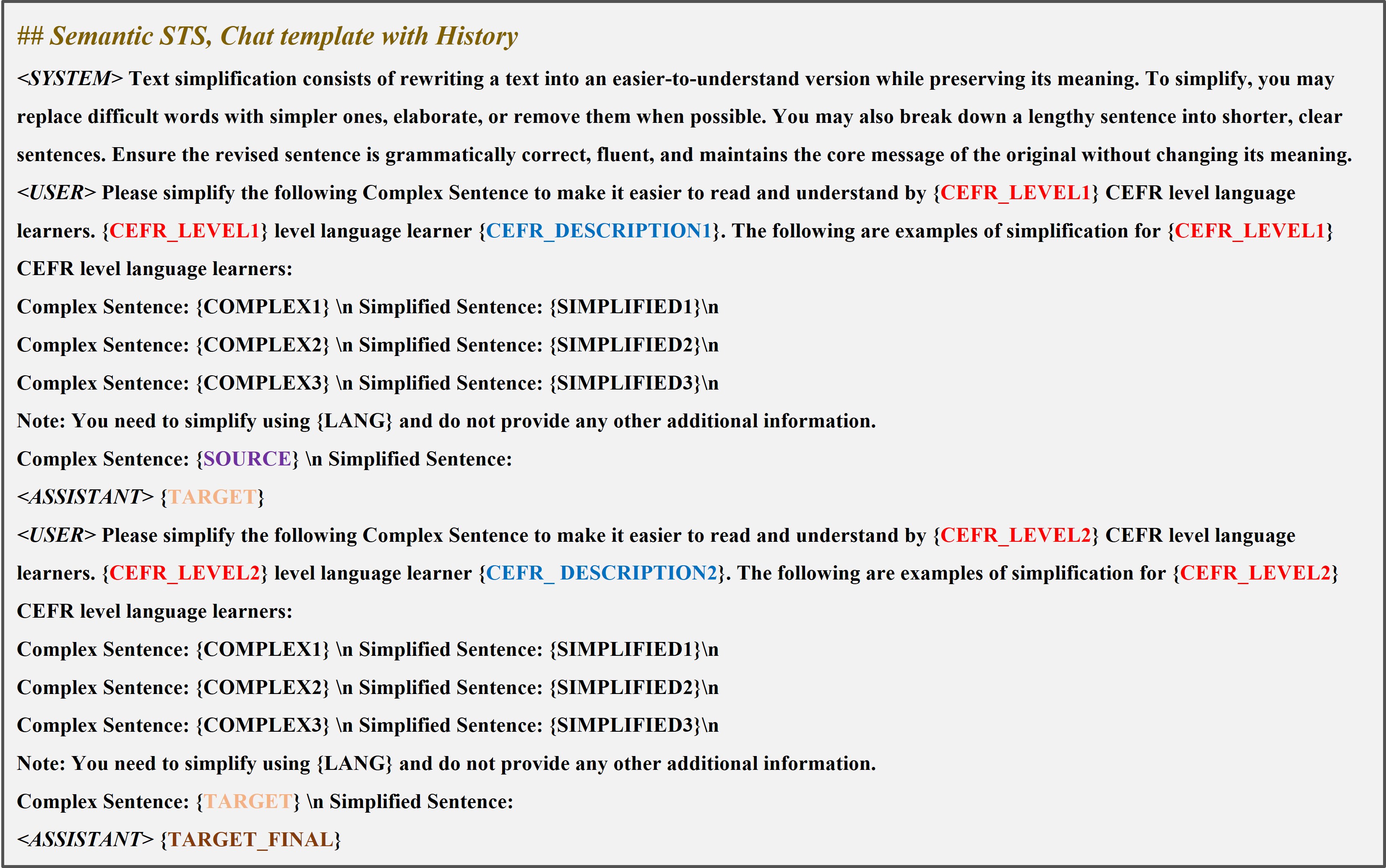}
\caption{Prompt Design for Few-shot CoT with Chat Template and Chat History}
\label{fig:chatHistory}
\end{figure*}

With the optimal path planning and top-k exemplar selection based on semantic preservation, we instruct LLMs to generate simplification step-by-step as illustrated in Figure~\ref{fig:framework}. \citet{barayan2025analysing} find that optimal performance is achieved through few-shot prompts using a single-case exemplar design that includes target, descriptor, and example. We enhance this approach by incorporating a chat template and chat history that includes the previous simplification steps along the optimal path. Figure~\ref{fig:chatHistory} shows our prompt design.

Given a test sentence at source proficiency level in $C_{src}$ with target level $C_{tgt}$, we first identify the optimal path $P^*$ using the reward matrix $R$. For each transition $(l_s, l_t)$ in $P^*$, we select exemplar prompt cases from $E_{s, t}$ to guide each step and instruct the LLM to generate a simplification for current transition. We used the generated simplified output $\hat{y}_t$ as the source sentence in the next iteration and append the prompt instruction from the previous step to the next prompt design to guide the LLM's CoT generation for the subsequent iteration. After reaching the final step toward the target level $C_{tgt}$, we use the sentence-level estimator $\phi(\cdot)$ to verify the CEFR level of the final output $\hat{y}_{tgt}$.

This approach uses validated exemplars to ensure meaning preservation during simplification. The step-by-step generation process  with chat history provides better control over the output quality.


\section{Experiment Design}

\subsection{Dataset}
We evaluate our approach on two benchmark datasets. The first is CEFR-SP (\citealp{arase2022cefr}), a sentence-level readability assessment corpus for English with CEFR annotations. CEFR-SP comprises three sources: Newsela-Auto (\citealp{xu2015problems}), Wiki-Auto (\citealp{jiang2020neural}), and SCoRe (\citealp{chujo2015corpus}). While \citet{barayan2025analysing} used only the publicly available portions (Wiki-Auto and SCoRe), we obtained additional access to the Newsela-Auto portion through direct contact with \citet{arase2022cefr}. We previously signed agreements with Newsela to use the dataset for research purposes only and will not redistribute it. The second public dataset is README++ (\citealp{naous2024readme}), which contains 9,757 sentences with manual CEFR annotations. These sentences span five languages (Arabic, English, French, Hindi, and Russian) and come from 112 different data sources to ensure domain diversity. We follow the standard train/dev/test splits provided in the official Github repositories of both datasets. The detailed data statistics are provided in Appendix~\ref{sec:dataStats} Table~\ref{tab:stats}.

\subsection{Sentence Level Estimator}

Our framework requires reliable sentence-level readability assessment for three key functions: reward matrix construction, exemplar selection, and output evaluation. We evaluated two state-of-the-art estimators: the CEFR-SP-estimator provided with the CEFR-SP dataset for English texts, and the README-estimator developed by \citealp{naous2024readme} which supports all five languages in their dataset.

To ensure optimal selection of sentence-level readability estimator, we conducted comprehensive benchmarking against various sentence readability assessment baselines, including traditional readability metrics and RSRS (\citealp{martinc2021supervised}) implementations with different language models. Our evaluations on both CEFR-SP and README++ datasets (Appendix ~\ref{sec:Estimator} (Tables~\ref{tab:cefr_classifier} and~\ref{tab:readme})) led us to adopt the CEFR-SP-estimator for CEFR-SP experiments and the README-estimator for README++ experiments, as they demonstrated superior performance on their respective datasets.

\subsection{Instruction-Tuned LLM}
We evaluated three open-source LLMs: Llama-3-8B-Instruct (\citealp{dubey2024llama}), Gemma-2-9b-it (\citealp{team2024gemma2}), and Mistral-7B-Instruct-v0.3 (\citealp{jiang2023mistral7b}). We find a consistent tradeoff pattern between simplification effectiveness and meaning preservation across all three LLMs. However, Mistral's performance is quite erratic. Therefore, we focus our results analysis on Llama and Gemma.

\subsection{Automatic Evaluation}
We use the following automatic metrics for simplification effectiveness and meaning preservation. \\
\textbf{Spearman Correlation ($\rho$)}: Rank correlation between target and estimated CEFR levels.\\
\textbf{RMSE}: Root mean squared error between estimated and target CEFR levels.\\
\textbf{Adjacent Accuracy}: Percentage of outputs within one CEFR level of the target.\\
\textbf{Exact Accuracy}: Percentage of output matching exactly the level of the target.\\
\textbf{Semantic Textual Similarity (STS)}: Cosine similarity between original and simplified texts, computed using sentence transformers (\citealp{reimers2019sentencebert}). We use all-MiniLM-L6-v2 for English and paraphrase-multilingual-MiniLM-L12-v2 for the other languages.\\
\textbf{BertScore}: Token-level semantic similarity between original and simplified texts, computed using BERT contextual embeddings (\citealp{zhang2020bertscore}).

\subsection{Human Evaluation}
\label{sec:human}
We further conduct comprehensive human evaluations on readability/proficiency and meaning preservation.\\
\textbf{Readability Assessment}
For each language, we recruited three evaluators, totaling 15 evaluators across five languages. All evaluators were non-native speakers with extensive university-level foreign language teaching experience. The English evaluators averaged 26 years of teaching English as a second language, French evaluators 13 years, Russian evaluators 6 years, Arabic evaluators 11 years, and Hindi evaluators 8 years.
To ensure familiarity with the CEFR descriptor scheme, we implemented a qualification task. For each language, we selected 30 sentences from benchmark datasets previously annotated with CEFR levels (10 each from levels B1, A2, and A1). We evaluated the annotators' CEFR assessments against gold references using three metrics: Majority Agreement, Krippendorff's Alpha, and Gwet AC2.

\textbf{Meaning Preservation Judgement}
For each language, the three instructor evaluators annotated 45 sentence pairs (original vs. simplified) across 9 source-target transitions for meaning preservation. Following \citet{barayan2025analysing}, we employed a scale where 0 indicates no or trivial changes; 1 signifies non-trivial changes with the main idea maintained; and 2 indicates failure to preserve the main idea. Intermediate scores (0.5 and 1.5) were also permitted.

From the 45 annotated pairs, we selected 20 pairs showing high agreement among instructor evaluators. High agreement was determined when at least two of our three metrics (Majority Agreement, Krippendorff's Alpha, and Gwet AC2) exceeded fair value thresholds. For these pairs, we used the majority judgment of the three annotators to establish meaning preservation scores, creating a qualification test for subsequent assessments.

We then recruited three native speakers for each language, all with college-level or higher education, to evaluate meaning preservation using these 20 sentence pairs. However, for French and Hindi, neither consistent agreement among the native speakers nor alignment with instructor reference scores was achieved. Consequently, for these two languages, we relied on the instructors for meaning preservation annotation. For English and Russian, only one native speaker per language achieved high agreement with instructor reference scores, so we paired these native speakers with two instructors to complete the annotation task. Similarly, for Arabic, we paired two native speakers with one instructor due to agreement.  We pay \$16 to \$18 per hour for human annotation. 

\subsection{Baselines}
We compare our research framework with the following baselines:\\
\textbf{COPY}: A simple baseline that returns the input sentence unchanged, serving as a lower bound for performance evaluation.\\
\textbf{Supervised}: An mT5-base model fine-tuned on available parallel simplification data for each language (details in Appendix~\ref{sec:Super}). For \textit{Supv. mT5-base (lang)}, mT5-base was fine-tuned on language-specific parallel data in the README dataset. For \textit{Supv. mT5-base (all)}, mT5-base was fine-tuned on combined parallel data available in the README dataset. Since CEFR dataset is English-only, we denote the supervised model as \textit{Supv. mT5-base} without further specification. \\
\textbf{\citet{barayan2025analysing} }: State-of-the-art results on the publicly available CEFR-SP dataset (Wiki-Auto and SCoRE portions), as reported in their paper.\\
\textbf{One-step single case}: Our reimplementation of \citet{barayan2025analysing} with one-step LLM generation and single-case prompt without semantic exemplars. We use their few-shot prompting without chat template. We run three sets of experiments with temperature of 1 and report average performances. Referencing the result Table~\ref{tab:readen_results} and~\ref{tab:all_results}, this model is configured as \textit{Llama\_w/o\_chat+N+N+N}. \\
\textbf{One-step single Chat}: We implement \citet{barayan2025analysing} with one-step LLM generation and prompt without semantic exemplars. But we use chat template without chat history. In the result Tables~\ref{tab:readen_results} and~\ref{tab:all_results}, this model is configured as \textit{Model+N+N+N}. \\
For our proposed framework, we experiment with the following models referencing notation in result Tables~\ref{tab:readen_results} and~\ref{tab:all_results}. \\
\textbf{One-step Semantic Chat} which is One-step LLM generation with semantic-preservation exemplars selection and chat template. The configuration is denoted as \textit{Model+N+Y+N}.\\
\textbf{Planning+Single-case CoT Chat History}: Two-stage approach using path planning and CoT generation with single-case exemplars and chat template and history. The configuration is denoted as \textit{Model+Y+N+Y}.\\
\textbf{Planning+Semantic CoT No History}: Two-stage approach using path planning and CoT generation with semantic exemplars and chat template without history. The configuration is denoted as \textit{Model+Y+Y+N}.\\
\textbf{Planning+Semantic Chat History}: Our complete framework combining path planning with semantic-aware CoT and chat history. The configuration is denoted as \textit{Model+Y+Y+Y}.

\section{Results and Analysis}
Due to space constraints, we present results on README-EN in the main text (Table~\ref{tab:readen_results}) and provide comprehensive results across all datasets in Appendix Table~\ref{tab:all_results}.

\begin{table*}[h]
\centering
\tiny
\setlength{\tabcolsep}{4pt}
\renewcommand{\arraystretch}{1.05}
\begin{tabular}{cccccccccc}
\hline
\textbf{Model} & \textbf{Planning} & \textbf{Semantic} & \textbf{History} & $\rho \uparrow$ & \textbf{AdjAcc} (\%) $\uparrow$ & \textbf{ExactAcc} (\%) $\uparrow$ & \textbf{RMSE} $\downarrow$ & \textbf{BS\_F1} $\uparrow$ & \textbf{STS} (\%) $\uparrow$ \\
\hline
COPY & -- & -- & -- & 0.00 & 32.32 & 6.82 & 2.29 & 100.0 & 100.0 \\
Supv. mT5-base (lang) & -- & --& --& 0.01 & 45.96 & 14.39 & 1.99 & 94.00 & 84.05 \\
Supv. mT5-base (all) & -- & -- & -- & 0.03 & 42.68 & 10.86 & 2.03 & 96.88 & 92.36 \\
Llama\_w/o\_chat & N & N & N & 0.12 & 62.62 & 24.66 & 1.60 & 91.05 & 70.41 \\
Llama & N & N & N & 0.45 & 72.98 & 26.26 & 1.27 & 90.72 & 73.43 \\
Llama & N & Y & N & 0.49 & 76.52 & 23.23 & 1.26 & 91.44 & 76.02 \\
Llama & Y & N & Y & 0.68 & 93.43 & 46.46 & 0.86 & 89.15 & 65.81 \\
Llama & Y & Y & N & 0.37 & 70.96 & 21.97 & 1.37 & 91.60 & 78.01 \\
Llama & Y & Y & Y & 0.66 & 91.41 & 37.37 & 0.95 & 90.00 & 69.90 \\
Gemma & N & N & N & 0.43 & 83.33 & 31.06 & 1.13 & 89.79 & 70.11 \\
Gemma & N & Y & N & 0.48 & 86.62 & 30.05 & 1.08 & 89.93 & 69.67 \\
Gemma & Y & N & Y & 0.66 & 96.46 & 44.70 & 0.81 & 88.10 & 61.32 \\
Gemma & Y & Y & N & 0.45 & 85.35 & 33.39 & 1.08 & 89.54 & 68.66 \\
Gemma & Y & Y & Y & 0.58 & 91.92 & 37.63 & 0.94 & 88.97 & 66.49 \\
\hline
\end{tabular}
\caption{Comprehensive results for README-EN.}
\label{tab:readen_results}
\end{table*}

\subsection{Trade-off Between Simplification Effectiveness and Meaning Preservation}
\label{sec:tradeoff}
Comparing Llama and Gemma models across all experiments, a clear trade-off pattern emerges. On README-EN, Gemma's Planning+Single case+CoT configuration achieves the highest adjacent accuracy (96.46\%) but lower STS (61.32\%), while Llama's same configuration reaches 93.43\% adjacent accuracy with higher STS (65.81\%). This inverse relationship between target level achievement and semantic preservation is consistent across both datasets. Llama is generally better at semantic preservation than Gemma but generally underperforms relative to Gemma in simplification. 

\subsection{Dynamic Path Planning Improves Readability Effectiveness}
Introducing dynamic path planning substantially boosts simplification effectiveness for both models. Comparing "Plan+single CoT" with "One-step single chat" configurations shows the impact of multi-step simplification. On README-EN, Llama's exact accuracy improves from 26.26\% (one-step single chat) to 46.46\% (Plan+single CoT with history), while Spearman correlation increases from 0.45 to 0.68. Similarly on CEFR-SP-Partial, Gemma improves from $\rho$ = 0.31 (one-step) to 0.59 (planning with history). This effectiveness gain comes with a trade-off in semantic preservation. For instance, on CEFR-SP-Partial with Llama, STS decreases from 79.56\% (one-step) to 71.46\% (planning), reflecting the inherent tension between aggressive simplification and meaning preservation. The inherent difficulty of large-span simplification is illustrated by representative examples in Appendix~\ref{sec:complex}.

\subsection{Chat History and Semantic Exemplars Optimize the Effectiveness-Preservation Balance}
Adding chat history which retains previous simplification steps further enhances level accuracy and stability. By making the LLM explicitly aware of its reasoning trajectory, chat history enables more coherent multi-step simplification. Across datasets, including history raises $\rho$ and adjacent accuracy by 3-6 points relative to the no-history variant. For example, on README-EN with Llama, "Plan+Semantic CoT with history" achieves $\rho$ = 0.66 and adjacent accuracy of 91.41\%, compared to $\rho$ = 0.37 and 70.96\% without history.

Semantic-aware exemplar selection provides complementary benefits for meaning preservation. Comparing "Plan+Semantic CoT" versus "Plan+single CoT" configurations reveals consistent improvements in semantic fidelity. On CEFR-SP-Partial with Llama, incorporating semantic exemplars improves STS from 71.46\% (single case) to 75.35\% (semantic selection), while maintaining strong readability control ($\rho$ = 0.56). This demonstrates that selecting exemplars based on meaning preservation helps guide the LLM toward simplifications that better retain semantic content.

Among all tested configurations, Planning + Semantic CoT + Chat History achieves the best overall balance—offering the highest readability control for Gemma and the strongest meaning preservation for Llama—demonstrating that dynamic, context-aware generation provides a robust framework for readability-controlled simplification.

\subsection{Human Evaluation Results}

\textbf{Inter-annotator reliability}. 
Table~\ref{tab:inter} shows acceptable inter-annotator agreement across all languages, with Krippendorff's Alpha ranging from 0.62-0.79 and Gwet AC2 from 0.5-0.68.

\begin{table*}[h]
	\centering
	\scriptsize
         \begin{tabular}{cccc}
		\toprule
		Language & Majority Agreement & Krippendorff's Alpha & Gwt AC2\\
		English & 91.11\% & 0.62 & 0.5 \\
		French & 93.33\% & 0.63 & 0.55\\
		Russian & 96.67\% & 0.73 & 0.67\\
		Arabic & 100\% & 0.7 & 0.59\\
		Hindi & 100\% & 0.79 & 0.68\\
		\bottomrule
	\end{tabular}
	\caption{Human Readability Assessment: Inter-annotator Agreement}
	\label{tab:inter}
\end{table*}

\noindent\textbf{Readability Control}. 
Table~\ref{tab:read} demonstrates strong simplification effectiveness of our Planning+Semantic CoT approach, achieving high Spearman correlation ($\rho$ = 0.49-0.66) and adjacent accuracy (95.45-100\%) across all languages.
\begin{table}[h]
	\centering
	\scriptsize
         \begin{tabular}{ccccc}
		\toprule
		Language & $\rho$ & Adj Acc (\%) & Exa Acc (\%) & RMSE\\
		English & 0.66 & 97.78 & 62.22 & 0.67 \\
		French & 0.64 & 97.78 & 60 & 0.68 \\
		Russian & 0.65 & 100 & 53.33 & 0.68\\
		Arabic & 0.63 & 97.78 & 60 & 0.68\\
		Hindi & 0.49 & 95.45 & 54.55 & 0.77\\
		\bottomrule
	\end{tabular}
	\caption{Human Readability Assessment for Planning+Semantic Chat History}
	\label{tab:read}
\end{table}

\textbf{Human Meaning Preservation Judgement}
Table~\ref{tab:meaning} reveals a critical finding: the trade-off between simplification effectiveness and meaning preservation observed in automatic evaluation (Section~\ref{sec:tradeoff}) is confirmed by human assessment. While one-step simplification maintains better semantic preservation (average scores 0.71-1.14), our Planning+Semantic CoT method shows higher meaning loss (0.93-1.47), particularly for large-span simplifications (C2 to A1).
Notably, establishing reliable meaning preservation judgments proved exceptionally challenging. Despite following established protocols (Section~\ref{sec:human}), we encountered unexpected difficulties in achieving annotator agreement. Surprisingly, native speakers showed even greater disagreement among themselves than language instructors. For French and Hindi, we ultimately relied on instructor judgments due to insufficient native speaker consensus. This annotation challenge underscores a fundamental difficulty: if human experts struggle to agree on meaning preservation, it highlights the inherent complexity LLMs face when attempting to maintain semantic fidelity during large-span simplifications.

\begin{table*}[ht]
\tiny
\centering
\caption{Human Judgement of Meaning Preservation}
\begin{tabular}{llcccccccccc}
\hline
\multirow{2}{*}{Language} & \multirow{2}{*}{Model} & \multirow{2}{*}{Avg. Score (Stdv)} & \multicolumn{3}{c}{C2 simplified to:} & \multicolumn{3}{c}{C1 simplified to:} & \multicolumn{3}{c}{B2 simplified to:} \\
\cmidrule(lr){4-6} \cmidrule(lr){7-9} \cmidrule(lr){10-12}
& & & B1 & A2 & A1 & B1 & A2 & A1 & B1 & A2 & A1 \\
\hline
\multirow{2}{*}{English} & Llama One-step single chat & 0.822 (0.327) & 0.6 & 0.6 & 1 & 0.6 & 0.4 & 0.9 & 0.7 & 1.2 & 1.4 \\
& Llama Plan+Semantic CoT (with hist.) & 1.089 (0.71) & 0.1 & 1.2 & 1.9 & 0.5 & 1.5 & 2 & 0.3 & 0.7 & 1.6 \\
\hline
\multirow{2}{*}{French} & Llama One-step single chat & 0.889 (0.513) & 0.5 & 1.5 & 1.3 & 0.1 & 0.4 & 0.6 & 0.9 & 1.5 & 1.2 \\
& Llama Plan+Semantic CoT (with hist.) & 0.933 (0.308) & 0.3 & 0.7 & 0.9 & 1 & 1.2 & 1.3 & 1 & 0.8 & 1.2 \\
\hline
\multirow{2}{*}{Russian} & Llama One-step single chat & 0.711 (0.257)& 0.5 & 0.8 & 1 & 0.6 & 0.6 & 0.9 & 0.6 & 1.1 & 0.3 \\
& Llama Plan+Semantic CoT (with hist.) & 1.133 (0.492) & 0.9 & 1.3 & 1.8 & 0.4 & 1.1 & 1.7 & 0.5 & 1.0 & 1.5 \\
\hline
\multirow{2}{*}{Arabic} & Llama One-step single chat & 1.144 (0.378) & 0.9 & 0.7 & 1 & 1.1 & 0.9 & 1.2 & 1.6 & 1.9 & 1.0 \\
& Llama Plan+Semantic CoT (with hist.) & 1.422 (0.441) & 0.6 & 1.3 & 2 & 0.9 & 1.4 & 1.8 & 1.5 & 1.6 & 1.7 \\
\hline
\multirow{2}{*}{Hindi} & Llama One-step single chat & 1.089 (0.386) & 1.0 & 0.9 & 1 & 1.3 & 1.1 & 1.3 & 1.9 & 0.6 & 0.7 \\
& Llama Plan+Semantic CoT (with hist.) & 1.467 (0.52) & 2.0 & 1.2 & 1.9 & 1.6 & 0.9 & 2.0 & 0.7 & 1.0 & 1.9 \\
\hline
\end{tabular}
\label{tab:meaning}
\end{table*}

\subsection{Efficiency of Path Planning}

To further evaluate the efficiency of dynamic path planning, we also perform an experiment where the simplification steps are sequential from source to target, with semantic CoT and chat history, to compare with Plan+Semantic CoT with Chat History. Evaluation shows that the simplification effectiveness scores and the meaning preservation scores are very similar between these two, indicating almost identical performances. Therefore, we examine the efficiency of path planning by measuring the ``reduction ratio of inference steps'' of Plan+Semantic CoT with Chat History compared to Sequential+Semantic CoT with Chat History. We present Llama results in Table~\ref{tab:pathllama} and Gemma results in Appendix Table~\ref{tab:pathgemma}. Results show that the dynamic path planning approach achieves a substantial reduction in computational overhead, decreasing the required processing steps by 22-42\%.

\begin{table}[h]
\tiny
        \centering
        \begin{tabular}{lcclc}
                \toprule
                \textbf{Corpus} & \textbf{Source} & \textbf{Target} & \textbf{DP-Path} & \textbf{RRIS$\uparrow$} \\
                \midrule
                \multirow{3}{*}{CEFR-SP-Partial} 
      & C2 & B1 & [B2, B1] & \multirow{3}{*}{25.0\%} \\
                 & C2 & A2 & [B2, B1, A2] &  \\
                 & C2 & A1 & [B2, B1, A2, A1] &  \\
                 \midrule
         \multirow{3}{*}{CEFR-SP-Whole} 
      & C2 & B1 & [B2, B1] & \multirow{3}{*}{25.0\%} \\
                 & C2 & A2 & [B2, B1, A2] &  \\
                 & C2 & A1 & [B2, B1, A2, A1] &  \\
                 \midrule
         \multirow{3}{*}{README\_EN} 
      & C2 & B1 & [B2, B1] & \multirow{3}{*}{25.0\%} \\
                 & C2 & A2 & [B2, B1, A2] &  \\
                 & C2 & A1 & [B2, B1, A2, A1] &  \\
                 \midrule
         \multirow{2}{*}{README\_FR} 
      & C2 & A2 & [B2, B1, A2] & \multirow{2}{*}{22.2\%} \\
                 & C2 & A1 & [B2, B1, A2, A1] &  \\
                 \midrule
        \multirow{3}{*}{README\_RU} 
              & C2 & B1 & [B2, B1] & \multirow{3}{*}{25.0\%} \\
                         & C2 & A2 & [B2, B1, A2] &  \\
                         & C2 & A1 & [B2, B1, A2, A1] &  \\
                \bottomrule
        \end{tabular}
        \caption{RRIS on Llama. RRIS denotes the \textbf{R}eduction \textbf{R}atio of \textbf{I}nference \textbf{S}teps relative to the Sequential-steps strategy.}
        \label{tab:pathllama}
\end{table}

\section{Conclusions}
We presented a novel framework for multilingual proficiency-controlled sentence simplification that addresses the fundamental challenge of simplifying across large readability spans. Our approach combines dynamic path planning, semantic-aware exemplar selection to preserve meaning, and chain-of-thought generation with chat history for coherent multi-step reasoning.
Comprehensive evaluation across five languages demonstrates that our framework significantly improves target level achievement compared to single-step approaches. Dynamic path planning also reduces computational steps by 22-42\% through efficient path optimization.
However, our results reveal an inherent trade-off: achieving precise readability control, especially for large-span simplifications, comes at the cost of semantic preservation. This challenge is validated by both automatic metrics and human evaluation, where even human annotators struggle to agree on meaning preservation judgments—highlighting the fundamental difficulty of this task. (see Appendix~\ref{sec:complex} for illustrative examples of the complexity gap between C2 and A1 levels). Future work should explore techniques to better balance this trade-off.

\section{Limitations}
While our approach shows promising results, several important limitations require attention. First, we focus on sentence-level simplification, while real-world applications also require document-level coherence and cross-sentence discourse relationships. Second, our human evaluation revealed fundamental challenges in achieving consensus on meaning preservation, suggesting that current evaluation paradigms may inadequately capture semantic fidelity. Third, the trade-off between readability control and meaning preservation intensifies with larger simplification spans, indicating that optimizing both dimensions simultaneously remains an open challenge. The writing of this paper was partially assisted by Claude for editing and refinement.



\section{Acknowledgments}
We thank the anonymous reviewers for their helpful comments and suggestions. 

\bibliography{customOct25}

\appendix

\section{Data Statistics}
\label{sec:dataStats}
We follow the standard train/dev/test splits from the benchmark datasets CEFR-SP and README++, as detailed in Table~\ref{tab:stats}. For both datasets, since the test sets are used to evaluate simplification from higher to lower levels, they only contain sentences from higher CEFR levels (B2, C1, and C2).

\begin{table*}[h]
    \centering
    \small
    \begin{tabular}{lcccccccc}
        \toprule
         & Split & A1 & A2 & B1 & B2 & C1 & C2 & Total \\
        \midrule
        \multirow{3}{*}{\textbf{\makecell{CEFR-SP \\ (Partial)}}} 
        & Train & 47 & 959 & 2245 & 2472 & 1364 & 91 & 7178 \\
        & Dev & 40 & 154 & 513 & 399 & 193 & 67 & 1366 \\
        & Test & - & - & - & 142 & 187 & 72 & 401 \\
        \midrule
        \multirow{3}{*}{\textbf{\makecell{CEFR-SP \\ (Whole)}}} 
        & Train & 55 & 1324 & 5481 & 5209 & 1821 & 100 & 13990 \\
        & Dev & 41 & 165 & 685 & 651 & 227 & 74 & 1843 \\
        & Test & - & - & - & 188 & 227 & 74 & 489 \\
        \midrule
        \multirow{3}{*}{\textbf{\makecell{README\_EN}}}
        & Train & 146 & 539 & 485 & 713 & 304 & 56 & 2243 \\
        & Dev & 14 & 69 & 61 & 91 & 39 & 9 & 283 \\
        & Test & - & - & - & 92 & 34 & 6 & 132 \\
        \midrule
        \multirow{3}{*}{\textbf{\makecell{README\_FR}}}
        & Train & 107 & 302 & 343 & 277 & 193 & 97 & 1319 \\
        & Dev & 14 & 31 & 52 & 29 & 22 & 17 & 165 \\
        & Test & - & - & - & 46 & 26 & 12 & 84 \\
        \midrule
        \multirow{3}{*}{\textbf{\makecell{README\_RU}}} 
        & Train & 321 & 223 & 334 & 260 & 193 & 73 & 1404 \\
        & Dev & 39 & 34 & 38 & 32 & 23 & 10 & 176 \\
        & Test & - & - & - & 34 & 21 & 8 & 63 \\
        \midrule
        \multirow{3}{*}{\textbf{\makecell{README\_AR}}} 
        & Train & 67 & 198 & 414 & 434 & 284 & 146 & 1543 \\
        & Dev & 6 & 26 & 44 & 63 & 38 & 19 & 196 \\
        & Test & - & - & - & 68 & 28 & 18 & 114 \\
        \midrule
        \multirow{3}{*}{\textbf{\makecell{README\_HI}}} 
        & Train & 204 & 230 & 223 & 200 & 172 & 150 & 1179 \\
        & Dev & 27 & 22 & 38 & 31 & 20 & 11 & 149 \\
        & Test & - & - & - & 32 & 30 & 13 & 75 \\
        \bottomrule
    \end{tabular}
    \caption{Data Statistics}
    \label{tab:stats}
\end{table*}

\section{Comprehensive Results Across All Datasets}
Table~\ref{tab:all_results} contains full experimental results for all language-dataset combinations, including detailes for each model configuration.

\begin{table*}[h]
\centering
\tiny
\setlength{\tabcolsep}{4pt}
\renewcommand{\arraystretch}{1.05}
\begin{tabular}{ccccccccccc}
\hline
\textbf{Dataset} & \textbf{Model} & \textbf{Planning} & \textbf{Semantic} & \textbf{History} & $\rho \uparrow$ & \textbf{AdjAcc} (\%) $\uparrow$ & \textbf{ExactAcc} (\%) $\uparrow$ & \textbf{RMSE} $\downarrow$ & \textbf{BS\_F1} $\uparrow$ & \textbf{STS} (\%) $\uparrow$ \\
\hline
\textbf{CEFR-SP-Part} & COPY  & -- & -- & -- & 0.00 & 24.15 & 3.99 & 2.45 & 100.0 & 100.0 \\
& Supv. mT5-base &-- & --& --& 0.01 & 39.43 & -- & 2.08 & 65.15 & 84.51 \\
& Barayan et al., 2025 &  N & N & N & 0.19 & 48.21 & -- & 1.80 & 55.28 & 82.86\\
& Llama\_w/o\_chat & N & N & N & 0.24 & 54.2 & 16.87 & 1.77 & 41.94 & 73.03 \\
& Llama & N & N & N  & 0.28 & 53.28 & 16.46 & 1.74 & 51.39 & 79.56 \\
& Llama & N & Y & N & 0.35 & 48.13 & 12.55 & 1.81 & 57.69 & 83.30 \\
& Llama & Y & N & Y & 0.52 & 74.73 & 30.01 & 1.35 & 39.66 & 71.46 \\
& Llama & Y & Y & N & 0.41 & 51.29 & 12.55 & 1.80 & 55.51 & 82.71 \\
& Llama & Y & Y & Y & 0.56 & 68.99 & 26.43 & 1.49 & 45.94 & 75.35 \\
& Gemma & N & N & N & 0.31 & 62.34 & 21.86 & 1.60 & 44.69 & 75.65 \\
& Gemma & N & Y & N & 0.40 & 63.67 & 20.45 & 1.56 & 46.60 & 76.55 \\
& Gemma & Y & N & Y & 0.59 & 83.87 & 38.99 & 1.13 & 32.36 & 65.66 \\
& Gemma & Y & Y & N  & 0.36 & 62.34 & 20.70 & 1.59 & 44.35 & 75.73 \\
& Gemma & Y & Y & Y & 0.49 & 72.07 & 28.10 & 1.42 & 39.28 & 72.77 \\
\hline
\textbf{CEFR-SP-Whole} & COPY & -- & -- & -- & 0.00 & 26.52 & 4.77 & 2.51 & 100.0 & 100.0 \\
& Supv. mT5-base &-- & --& -- & 0.00 & 38.85 & 11.52 & 2.20 & 89.17 & 70.57 \\
& Llama\_w/o\_chat & N & N & N  & 0.06 & 47.31 & 16.54 & 2.05 & 58.77 & 78.03 \\
& Llama & N & N & N  & 0.29 & 55.08 & 17.45 & 1.69 & 51.90 & 79.49 \\
& Llama & N & Y & N & 0.35 & 50.72 & 13.84 & 1.76 & 57.68 & 82.62 \\
& Llama & Y & N & Y & 0.53 & 77.51 & 31.42 & 1.29 & 40.04 & 71.11 \\
& Llama & Y & Y & N & 0.37 & 52.01 & 13.77 & 1.78 & 56.57 & 83.06 \\
& Llama & Y & Y & Y & 0.55 & 70.69 & 27.33 & 1.46 & 47.12 & 75.68 \\
& Gemma & N & N & N  & 0.32 & 64.62 & 23.31 & 1.54 & 45.19 & 75.25 \\
& Gemma & N & Y & N & 0.36 & 67.01 & 24.40 & 1.51 & 45.29 & 75.14 \\
& Gemma & Y & N & Y & 0.19 & 50.65 & 16.56 & 1.85 & 32.70 & 65.53 \\
& Gemma & Y & Y & N & 0.36 & 66.39 & 24.06 & 1.49 & 41.97 & 72.40 \\
& Gemma & Y & Y & Y & 0.51 & 78.05 & 32.31 & 1.28 & 37.06 & 69.34 \\
\hline
\textbf{README-AR} & COPY & -- & -- & -- & 0.00 & 26.02 & 4.68 & 2.61 & 100.0 & 100.0 \\
& Supv. mT5-base (all) & -- & -- & -- & 0.01 & 55.56 & 21.35 & 1.95 & 90.63 & 75.39 \\
& Llama\_w/o\_chat & N & N & N & 0.11 & 44.93 & 9.26 & 1.91 & 85.56 & 66.37 \\
& Llama & N & N & N & 0.18 & 48.54 & 15.50 & 1.94 & 91.79 & 81.61 \\
& Llama & N & Y & N  & 0.24 & 48.25 & 15.20 & 1.95 & 92.59 & 83.81 \\
& Llama & Y & N & Y & 0.53 & 73.39 & 20.47 & 1.39 & 89.56 & 71.27 \\
& Llama & Y & Y & N & 0.27 & 54.09 & 17.54 & 1.76 & 91.86 & 80.22 \\
& Llama & Y & Y & Y & 0.58 & 74.27 & 17.54 & 1.39 & 90.70 & 74.38 \\
& Gemma & N & N & N & 0.25 & 55.85 & 18.71 & 1.73 & 89.90 & 74.89 \\
& Gemma & N & Y & N & 0.26 & 59.06 & 18.42 & 1.69 & 90.08 & 75.61 \\
& Gemma & Y & N & Y & 0.51 & 78.65 & 24.56 & 1.33 & 88.30 & 66.96 \\
& Gemma & Y & Y & N & 0.28 & 62.57 & 21.05 & 1.59 & 89.77 & 74.49 \\
& Gemma & Y & Y & Y & 0.45 & 69.30 & 20.76 & 1.50 & 89.38 & 72.27 \\
\hline
\textbf{README-EN} & COPY & -- & -- & -- & 0.00 & 32.32 & 6.82 & 2.29 & 100.0 & 100.0 \\
& Supv. mT5-base (lang) & -- & --& --& 0.01 & 45.96 & 14.39 & 1.99 & 94.00 & 84.05 \\
& Supv. mT5-base (all) & -- & -- & -- & 0.03 & 42.68 & 10.86 & 2.03 & 96.88 & 92.36 \\
& Llama\_w/o\_chat & N & N & N & 0.12 & 62.62 & 24.66 & 1.60 & 91.05 & 70.41 \\
& Llama & N & N & N & 0.45 & 72.98 & 26.26 & 1.27 & 90.72 & 73.43 \\
& Llama & N & Y & N & 0.49 & 76.52 & 23.23 & 1.26 & 91.44 & 76.02 \\
& Llama & Y & N & Y & 0.68 & 93.43 & 46.46 & 0.86 & 89.15 & 65.81 \\
& Llama & Y & Y & N & 0.37 & 70.96 & 21.97 & 1.37 & 91.60 & 78.01 \\
& Llama & Y & Y & Y & 0.66 & 91.41 & 37.37 & 0.95 & 90.00 & 69.90 \\
& Gemma & N & N & N & 0.43 & 83.33 & 31.06 & 1.13 & 89.79 & 70.11 \\
& Gemma & N & Y & N & 0.48 & 86.62 & 30.05 & 1.08 & 89.93 & 69.67 \\
& Gemma & Y & N & Y & 0.66 & 96.46 & 44.70 & 0.81 & 88.10 & 61.32 \\
& Gemma & Y & Y & N & 0.45 & 85.35 & 33.39 & 1.08 & 89.54 & 68.66 \\
& Gemma & Y & Y & Y & 0.58 & 91.92 & 37.63 & 0.94 & 88.97 & 66.49 \\
\hline
\textbf{README-FR} & COPY  & -- & -- & --& 0.00 & 28.57 & 6.75 & 2.42 & 100.0 & 100.0 \\
& Supv. mT5-base (lang) & -- & --& -- & 0.12 & 75.79 & 21.83 & 1.39 & 84.16 & 43.76 \\
& Supv. mT5-base (all) & -- & -- & -- & 0.05 & 44.44 & 14.29 & 1.99 & 95.56 & 88.72 \\
& Llama\_w/o\_chat & N & N & N  & 0.22 & 57.28 & 19.44 & 1.66 & 89.55 & 76.93 \\
& Llama & N & N & N  & 0.28 & 48.81 & 14.29 & 1.79 & 93.22 & 85.77 \\
& Llama & N & Y & N & 0.31 & 51.59 & 12.70 & 1.76 & 93.45 & 86.08 \\
& Llama & Y & N & Y  & 0.67 & 79.76 & 20.63 & 1.21 & 91.03 & 76.38 \\
& Llama & Y & Y & N  & 0.38 & 51.59 & 13.10 & 1.70 & 92.91 & 84.77 \\
& Llama & Y & Y & Y & 0.68 & 76.98 & 17.46 & 1.26 & 91.75 & 78.23 \\
& Gemma & N & N & N & 0.53 & 63.49 & 18.65 & 1.43 & 91.71 & 80.74 \\
& Gemma & N & Y & N  & 0.50 & 66.27 & 21.03 & 1.38 & 91.77 & 80.19 \\
& Gemma & Y & N & Y & 0.75 & 89.68 & 27.78 & 1.03 & 89.69 & 69.35 \\
& Gemma & Y & Y & N & 0.52 & 69.84 & 21.83 & 1.36 & 91.14 & 77.14 \\
& Gemma & Y & Y & Y & 0.66 & 81.75 & 25.79 & 1.14 & 90.65 & 74.93 \\
\hline
\textbf{README-RU} & COPY & -- & -- & -- & 0.00 & 18.52 & 4.23 & 2.73 & 100.0 & 100.0 \\
& Supv. mT5-base (lang) & -- & --& -- & 0.00 & 64.02 & 23.28 & 1.60 & 92.33 & 79.77 \\
& Supv. mT5-base (all) & -- & -- & -- & 0.08 & 52.38 & 17.46 & 1.91 & 94.18 & 87.25 \\
& Llama\_w/o\_chat & N & N & N  & 0.29 & 55.38 & 19.75 & 1.83 & 89.62 & 75.90 \\
& Llama & N & N & N  & 0.34 & 68.78 & 23.81 & 1.45 & 92.60 & 84.73 \\
& Llama & N & Y & N  & 0.43 & 68.25 & 17.99 & 1.48 & 93.09 & 86.20 \\%
& Llama & Y & N & Y & 0.76 & 97.35 & 59.26 & 0.70 & 90.36 & 73.59 \\
& Llama & Y & Y & N & 0.41 & 67.72 & 20.63 & 1.42 & 92.47 & 83.65 \\
& Llama & Y & Y & Y & 0.73 & 93.12 & 49.74 & 0.84 & 91.15 & 76.19 \\
& Gemma & N & N & N & 0.53 & 85.71 & 41.80 & 1.03 & 91.12 & 78.35 \\
& Gemma & N & Y & N & 0.51 & 84.13 & 32.28 & 1.16 & 91.75 & 81.14 \\
& Gemma & Y & N & Y & 0.77 & 99.47 & 64.02 & 0.61 & 89.65 & 69.01 \\
& Gemma & Y & Y & N & 0.64 & 92.06 & 45.50 & 0.90 & 91.18 & 78.35 \\
& Gemma & Y & Y & Y & 0.70 & 93.12 & 50.26 & 0.84 & 90.72 & 75.06 \\
\hline
\textbf{README-HI} & COPY & -- & -- & -- & 0.00 & 28.00 & 8.00 & 3.09 & 100.0 & 100.0 \\
& Supv. mT5-base (all) & -- & -- & --  & 0.01 & 51.11 & 14.67 & 2.24 & 93.04 & 75.19 \\
& Llama\_w/o\_chat & N & N & N & 0.13 & 48.74 & 18.66 & 2.30 & 89.98 & 74.63 \\
& Llama & N & N & N  & 0.21 & 45.33 & 12.44 & 2.30 & 94.05 & 86.07 \\
& Llama & N & Y & N & 0.20 & 44.44 & 12.44 & 2.34 & 94.87 & 88.50 \\
& Llama & Y & N & Y & 0.62 & 80.89 & 41.78 & 1.30 & 91.84 & 73.75 \\
& Llama & Y & Y & N & 0.25 & 51.11 & 17.78 & 1.99 & 93.36 & 83.16 \\
& Llama & Y & Y & Y & 0.53 & 68.44 & 30.22 & 1.60 & 92.90 & 78.27 \\
& Gemma & N & N & N & 0.34 & 68.44 & 25.78 & 1.55 & 92.19 & 77.22 \\
& Gemma & N & Y & N & 0.32 & 67.56 & 24.89 & 1.62 & 92.60 & 79.23 \\
& Gemma & Y & N & Y & 0.48 & 82.67 & 39.56 & 1.24 & 91.07 & 70.98 \\
& Gemma & Y & Y & N & 0.33 & 68.89 & 28.44 & 1.62 & 92.19 & 77.59 \\
& Gemma & Y & Y & Y & 0.40 & 72.89 & 30.67 & 1.47 & 91.94 & 75.80 \\
\hline
\end{tabular}
\caption{Comprehensive results for All Datasets.}
\label{tab:all_results}
\end{table*}

\section{Comparing Sentence Level Estimators}
\label{sec:Estimator}
\subsection{CEFR-SP}
We conducted extensive comparisons of readability assessment methods across both datasets. Table~\ref{tab:cefr_classifier} presents the evaluation of different approaches on CEFR-SP data, comparing unsupervised models (RSRS trained with BERT and XLM-R variants), traditional metrics (FKGL, ARI), and CEFR-SP-estimator and README-estimator. Traditional metrics showed limited correlation (FKGL: 0.734, ARI: 0.71), while neural RSRS \cite{martinc2021supervised} implementations achieved moderate performance (RSRS\_XLMR\_B: 0.678). The CEFR-SP estimator demonstrated the strongest performance (0.937).
\begin{table*}[h!]
    \centering
    \small
    \begin{tabular}{ccccccccc}
    \toprule
    Model & CEFR-SP (Partial) & CEFR-SP (Whole) \\
    \midrule
    CEFR-SP-estimator & 0.937 & 0.933  \\
    \midrule
    README-estimator\_EN & 0.817 & 0.797  \\
    \midrule
    RSRS\_BERT\_L & 0.535 & 0.515  \\
    \midrule
    RSRS\_BERT\_B & 0.552 & 0.534  \\
    \midrule
    RSRS\_XLMR\_L & 0.67 & 0.645  \\
    \midrule
    RSRS\_XLMR\_B & 0.678 & 0.654  \\
    \midrule
    RSRS\_mBERT & 0.627 & 0.609  \\
    \midrule
    RSRS\_mT5\_L & 0.417 & 0.4  \\
    \midrule
    RSRS\_mT5\_L & 0.611 & 0.585  \\
    \midrule
    RSRS\_mT5\_L & 0.507 & 0.478  \\
    \midrule
    FKGL & 0.734 & 0.705  \\
    \midrule
    ARI & 0.71 & 0.68  \\
    \midrule
    SL & 0.659 & 0.627  \\
    \bottomrule
    \end{tabular}
    \caption{Comparing CEFR-SP-estimator, README-estimator, with unsupervised models and traditional metrics on CEFR-SP datasets. Performance measured with Spearman's Correlation \textbf{$\rho$ ($\uparrow$)}}
        \label{tab:cefr_classifier}
\end{table*}

\subsection{README}
Table~\ref{tab:readme} compares README estimator and CEFR estimator on the README dataset. Although README-estimator achieves 0.00 F1 on category C2, most of its predictions fall into the adjacent category C1, resulting in minimal practical error. The consistent superior F1 over all levels except for C2 and the higher Average F1 score (49.59 vs. 47.9) justify selecting README-estimator over CEFR-SP-estimator.
\begin{table*}[h!]
    \centering
    \small
    \setlength{\tabcolsep}{4pt} 
    \begin{tabular}{ccccccccc}
    \toprule
    Model & A1 & A2 & B1 & B2 & C1 & C2 & Average F1 \\
    \midrule
    CEFR-SP-estimator & 50.00 & 45.10 & 50.56 & 56.41 & 56.76 & 28.57 & 47.9 \\
    \midrule
    README-estimator & 55.81 & 57.97 & 59.52 & 66.67 & 57.58 & 0.00 & 49.59 \\
    \bottomrule
    \end{tabular}
    \caption{Comparing  CEFR-SP-estimator and README-estimator on README\_EN dataset. Performance measured with F1. }
      \label{tab:readme}
\end{table*}

\section{Model Parameters}
\subsection{LLM Inference Setting}
We conduct the inference process on an NVIDIA GeForce RTX 3080, leveraging the Transformers library (\citealp{wolf2020transformers}). For generation, we set a maximum output length of 128 tokens, a temperature of 0.0 and do\_sample=False to ensure reproducibility.

\subsection{Supervised mt5 Training and Inference Setting}
\label{sec:Super}
For the supervised baseline model training, the batch size was 4, and the optimizer used was Adam (\citealp{kingma2017adam}) with learning rate of 2e-5. For $Supv. mT5-base (lang)$, mT5-base was fine-tuned on language-specific parallel data only in the README dataset. Since CEFR dataset is English-only, we denote the supervised model as $Supv. mT5-base$ without further specification. We configure the validation step to be 500 and conducted the validation process 10 times. As for $Supv. mT5-base (all)$, mT5-base was fine-tuned on combined parallel data available in the READMe dataset.  the validation step was set to 1500, also with 10 repetitions of the validation. Subsequently, we adopted SARI (\citealp{xu2016optimizing}) and selected the checkpoint achieving the highest SARI score on the validation set as the final model. The total GPU computation time was under 4 hours for mt5 training and inference.

\section{Supervised Baseline Training and Inference}
For the supervised baseline experiments, we constructed parallel datasets for different languages following \citet{ryan2023revisiting}:\\
English: We followed the methodology from \citet{barayan2025analysing}, though our train/dev split distribution differs slightly as the original split details were not available.\\
French: We combined data from two sources: WikiLarge (\citealp{cardon2020french}, randomly selected 5,000 sentence pairs) and CLEAR (\citealp{grabar2018clear}, included all available sentence pairs). The distribution of this data across CEFR levels is shown in Table 5.\\
Russian: Utilized the complete RSSE (\citealp{sakhovskiy2021rusimplesenteval}) dataset.\\
For Hindi and Arabic, no publicly available parallel sentence datasets were available for text simplification.\\
We implemented two training strategies. The first approach trained individual models for each language and generated simplifications on the validation set. The second strategy fine-tuned mT5-base using combined English, French, and Russian data, then generated simplifications on validation sets.
For implementation, we used mT5-base as our base model. All sentences were prepended with their corresponding CEFR level during training and validation. The parallel sentence pairs consisted of source and simplified versions, with validation performed on language-specific validation sets. Table~\ref{tab:sent} presents the detailed distribution of training and validation data across CEFR levels.

\begin{table*}[h]
    \centering
    \small
    \begin{tabular}{ccccccc}
        \toprule
        - & A1 & A2 & B1 & B2 & C1 & Total \\
        \midrule
        \multicolumn{7}{l}{English:} \\
        Train & 96 & 458 & 1277 & 206 & 10 & 2045 \\
        Val & 26 & 133 & 303 & 47 & 3 & 512 \\
        \midrule
        \multicolumn{7}{l}{French:} \\
        Train & 0 & 908 & 700 & 345 & 23 & 1976 \\
        Val & 0 & 220 & 179 & 93 & 3 & 495 \\
        \midrule
        \multicolumn{7}{l}{Russian:} \\
        Train & 428 & 772 & 368 & 72 & 0 & 1640 \\
        Val & 117 & 182 & 86 & 26 & 0 & 411 \\
        \bottomrule
    \end{tabular}
    \caption{Distribution of sentence target levels in training and validation sets for Supervised models}
    \label{tab:sent}
\end{table*}

\section{Human Evaluation Instruction Screenshots}
\begin{figure}
\centering
\includegraphics[width=\linewidth]{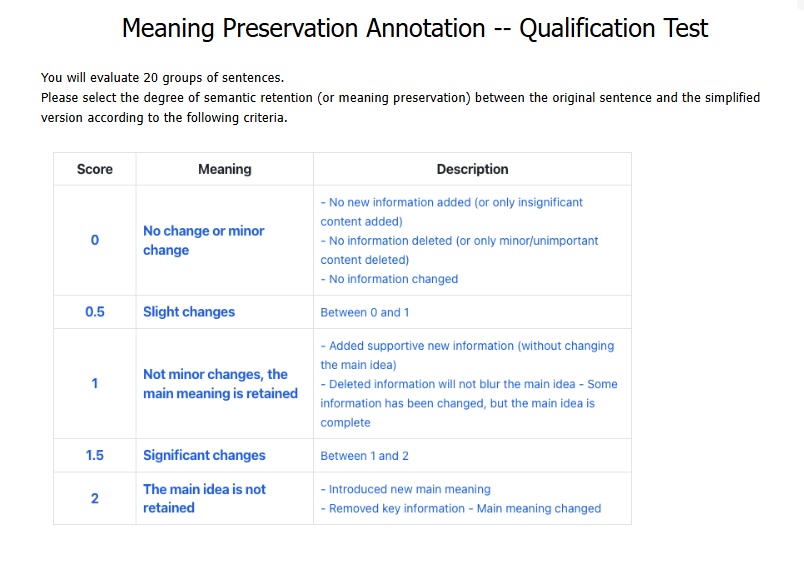}
\caption{Screenshot of Human Meaning Preservation Judgement: Instruction}
\label{fig:instruct}
\end{figure}

\begin{figure}
\centering
\includegraphics[width=\linewidth]{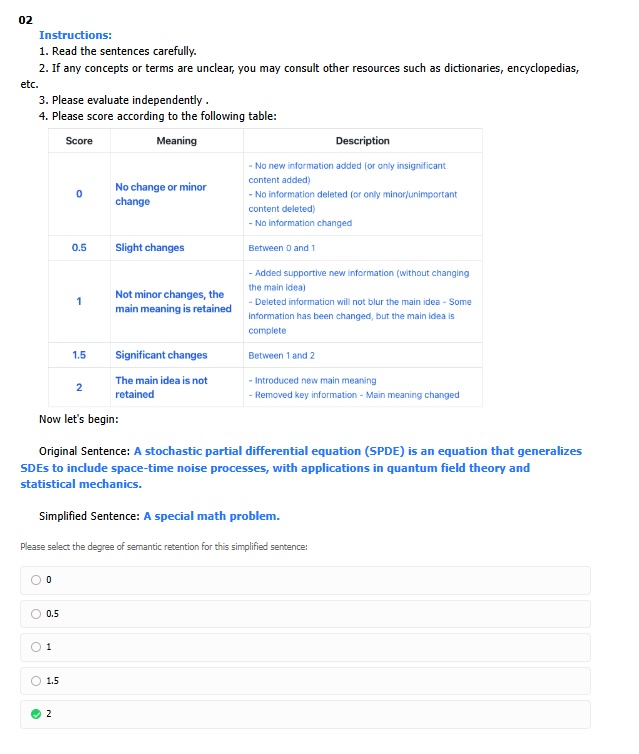}
\caption{Screenshot of Human Meaning Preservation Judgement: Annotation}
\label{fig:anno}
\end{figure}

Figures~\ref{fig:instruct} and ~\ref{fig:anno} provide screenshots of the user interface and detailed instructions given to human annotators for both readability assessment and meaning preservation tasks.

\section{Efficiency of Path Planning}
Table~\ref{tab:pathgemma} compares dynamic path planning against sequential simplification based on Gemma, including inference step reduction ratios across different languages.
\begin{table*}[h]
\small
        \centering
        \begin{tabular}{lcclc}
                \toprule
                \textbf{Corpus} & \textbf{Source} & \textbf{Target} & \textbf{DP-Path} & \textbf{RRIS$\uparrow$} \\
                \midrule
                \multirow{3}{*}{CEFR-SP-Partial} 
      & C2 & B1 & [B2, B1] & \multirow{3}{*}{25.0\%} \\
                 & C2 & A2 & [B2, B1, A2] &  \\
                 & C2 & A1 & [B2, B1, A2, A1] &  \\
                 \midrule
         \multirow{3}{*}{CEFR-SP-Whole} 
      & C2 & B1 & [B2, B1] & \multirow{3}{*}{25.0\%} \\
                 & C2 & A2 & [B2, B1, A2] &  \\
                 & C2 & A1 & [B2, B1, A2, A1] &  \\
                 \midrule
         \multirow{3}{*}{README\_EN} 
      & C2 & B1 & [B2, B1] & \multirow{3}{*}{25.0\%} \\
                 & C2 & A2 & [B2, B1, A2] &  \\
                 & C2 & A1 & [B2, B1, A2, A1] &  \\
                 \midrule
         \multirow{2}{*}{README\_AR} 
      & C2 & B1 & [B2, B1] & \multirow{2}{*}{28.57\%} \\
                 & C2 & A2 & [B2, B1, A2] &  \\
                 \midrule
         \multirow{2}{*}{README\_FR} 
      & C2 & A2 & [B2, B1, A2] & \multirow{2}{*}{22.2\%} \\
                 & C2 & A1 & [B2, B1, A2, A1] &  \\
                 \midrule
        \multirow{6}{*}{README\_RU} 
              & C2 & B1 & [B2, B1] & \multirow{6}{*}{27.27\%} \\
                         & C2 & A2 & [B2, B1, A2] &  \\
                         & C2 & A1 & [B2, B1, A2, A1] &  \\
              & C1 & A2 & [B1, A2] &  \\
              & C1 & A1 & [B1, A2, A1] &  \\
              & B2 & A1 & [A2, A1] &  \\
    \midrule
    \multirow{5}{*}{README\_HI} 
              & C2 & A2 & [B1, A2] & \multirow{5}{*}{42.11\%} \\
                         & C2 & A1 & [B1, A2, A1] &  \\
              & C1 & A2 & [B1, A2] &  \\
              & C1 & A1 & [B1, A2, A1] &  \\
              & B2 & A1 & [A1] &  \\
                \bottomrule
        \end{tabular}
        \caption{RRIS on Gemma. RRIS denotes the \textbf{R}eduction \textbf{R}atio of \textbf{I}nference \textbf{S}teps relative to the Sequential-steps strategy.}
                \label{tab:pathgemma}
\end{table*}

\section{Illustrative Examples of CEFR Level Complexity}
\label{sec:complex}
To illustrate the inherent difficulty of large-span proficiency-controlled simplification, we present representative examples from our benchmark datasets showing the stark contrast between beginner (A1) and expert (C2) level texts.

\textbf{A1 Level Examples}:
\begin{itemize}
\item "How was your first day in college?"
\item "That's a good question."
\item "And I can tell you this, this can be done."
\end{itemize}

A1 represents the beginner level where sentences are short, use common everyday vocabulary, simple grammatical structures, and express concrete, straightforward ideas.

\textbf{C2 Level Examples:}

\textit{Technical domain:}

"This is referred to as the dual space-time variation property and it indicates that precoding using arbitrary joint space-time orthogonal basis is not sufficient to ensure interference-free communication, unless these basis remain orthogonal after propagating through the channel."

\textit{Literary domain:}

"Would you allow me, in my gratitude for the benevolent reception that you gave me one day, to draw the attention of your rightful glory and to tell you that your star, so happy until now, is threatened by the most shameful and most ineffaceable of blemishes?"

C2 represents mastery level where sentences contain domain-specific terminology, complex syntactic structures, abstract concepts, and sophisticated lexical choices.

\textbf{The Inherent Challenge:}
These examples illustrate why simplification from C2 to A1 entails inevitable trade-offs.  Transforming such content to beginner-level comprehensibility requires compression, abstraction, and semantic loss. Our framework does not eliminate this fundamental trade-off but rather makes it controllable and optimized through multi-step planning and semantic-aware generation.

\end{document}